\documentclass[letterpaper, 10 pt, conference]{ieeeconf}
\IEEEoverridecommandlockouts
\usepackage{moresize}
\usepackage{graphicx}
\usepackage{psfrag}
\usepackage{amsmath}
\usepackage{algorithm,algpseudocode}
\usepackage{amssymb}
\usepackage{graphicx}
\usepackage{subfigure}
\usepackage{verbatim}
\usepackage{latexsym}
\usepackage[dvipsnames]{xcolor}
\usepackage{cite}
\usepackage{stmaryrd}
\usepackage{pifont}
\usepackage{textcomp}
\usepackage{gensymb}
\usepackage{colortbl}
\usepackage{multirow}
\usepackage{caption}
\usepackage{hyperref}
\usepackage{epstopdf}
\usepackage{cleveref}
\usepackage{tabto}
\usepackage{tikz}
\graphicspath{./images}
\DeclareGraphicsExtensions{.pdf,.eps, .jpg, .png, .PNG}
\title{\LARGE \bf Control Framework for a Hybrid-steel Bridge Inspection Robot}
\vspace{-15pt}
\author{Hoang-Dung Bui, Son  Nguyen, U-H. Billah, Chuong Le, Alireza Tavakkoli, Hung M. La, \textit{IEEE Senior Member}
\thanks{This work is supported by the U.S. National Science Foundation (NSF) under grants NSF-CAREER: 1846513 and NSF-PFI-TT: 1919127, and the U.S. Department of Transportation, Office of the Assistant Secretary for Research and Technology (USDOT/OST-R) under Grant No. 69A3551747126 through INSPIRE University Transportation Center, the Vingroup Innovation Foundation (VINIF) in project code VINIF.2020.NCUD.DA094, and the Japan NineSigma through the Penta-Ocean Construction Ltd. Co. under Agreement No. SP-1800087. The views, opinions, findings and conclusions reflected in this publication are solely those of the authors and do not represent the official policy or position of the NSF, the USDOT/OST-R and any other entities.
}
\thanks{The authors are with the Advanced Robotics and Automation (ARA) Lab, Department of Computer Science and Engineering, University of Nevada, Reno, NV  89557, USA.  Corresponding author: Hung La, email: hla@unr.edu.}
}

\begin{document}
\vspace{-15pt}
\maketitle

\begin{abstract} 
Autonomous navigation of steel bridge inspection robots are essential for proper maintenance. Majority of existing robotic solutions for bridge inspection require human intervention to assist in the control and navigation. In this paper, a control system framework has been proposed for a previously designed ARA robot \cite{icra20}, which facilitates autonomous real-time navigation and minimizes human involvement. The mechanical design and control framework of ARA robot enables two different configurations, namely the mobile and inch-worm transformation. In addition, a switching control was developed  with 3D point clouds of steel surfaces as the input which allow the robot to switch between mobile and inch-worm transformation. The surface availability algorithm (considers plane, area and height) of the switching control enables the robot to perform inch-worm jumps autonomously. The \textit{mobile transformation} allows the robot to move on continuous steel surfaces and perform visual inspection of steel bridge structures. Practical experiments on actual steel bridge structures highlight the effective performance of ARA robot with the proposed control framework for autonomous navigation during visual inspection of steel bridges. 
\end{abstract}

\section{Introduction}\label{S.1}
Within the field of health monitoring of bridge structures, the development of novel robotic platforms has received considerable attention in the recent years. It has been increasingly stressed in the literature that timely and regular monitoring of steel bridges ensures the safety of transportation vehicles. Environmental degradation (e.g. rain, wind, solar radiation), continuous surface-level friction, overloading, and other factors lead to deterioration of different structures on steel bridges. Continuous steel bridge monitoring is necessary to ensure transportation safety and proper maintenance. Most of these bridges are monitored by civil inspectors manually~\cite{NguyenLa_IROS2019}. However, due to the complex structural composition and inaccessible regions of the bridges (e.g. pipes, poles, overhead cables), the manual inspection of these regions is a perilous task for human inspector. Additionally, manual inspection is time consuming, labor-intensive, and disruptive to traffic. It is for this reason that different robotic solutions have been developed for automated steel bridge inspection \cite{Pham_ICERA2020,Nguyen_SHMII2019, Seo_TMECH2013, La_Robotica_2019, Wang_TRO2017, PL_Allerton2016, Kamdar2015, PL_ISARC2016, MaggHD}. These robots were equipped with different adhesion mechanism (e.g. magnetic wheels, pneumatic, suction cups, bio-inspired grippers), visual sensors (e.g. monocular, stereovision, RGB-D sensors) and other sensory modalities to facilitate navigation and inspection (e.g. IMUs, eddy current sensors)~\cite{Pham_ICERA2020,La_Robotica_2019, Wang_TRO2017, Kamdar2015, PL_Allerton2016,PL_ISARC2016, Seo_TMECH2013}. Magnetic adhesion enabled MaggHD robot \cite{MaggHD} to navigate flexibly on smooth steel surfaces. The incorporation of legged mechanism with electromagnets allow robots to assist in locomotion and traversal through complex steel structures \cite{Magnapods}. These robots are designed for a particular environment, lacking the deploy-ability in many unstructured environments. A flexible and versatile climbing robot was designed in \cite{NguyenLa_IROS2019}, which was equipped with 5-DOF arm, eddy current sensor and RGB-D sensors for inspection of steel bridge surfaces, especially for inaccessible regions of the bridges. Another type of climbing robot was developed by \cite{Versatrax} with untouched magnet blocks to move efficiently on metal surfaces. Although these robots alleviated the difficulty of moving on complex steel surfaces, they were controlled manually by cables or remote instruction from human operators.

As an effort to navigate robots autonomously, a hybrid robot design (named as ARA robot) was proposed by our previous research \cite{icra20}, which can travel on a smooth steel surface in \textit{mobile transformation} and transit from one steel surface to another in \textit{inch-worm transformation}. Since steel bridge inspection is a continuous process, the primary goal of our research is to develop a fully autonomous robotic system to automate this task. In this paper, we propose a control framework for ARA robot to navigate autonomously on steel bridge structures. A switching control mechanism is developed to allow the robot to determine availability of planar surface, which also facilitates the robot to perform in two different transformations. The switching control determines the availability of the planar surface, its area and height for determining its next transition. An area estimation algorithm has been proposed using point cloud data from RGB-D sensor, which allows the robot to assess area availability for transitioning from one plane to another. This algorithm determines if the available area is sufficient enough for the robots foot transition. Based on the height estimation on switching control, the robot chooses its transformation. The robot performs an inch-worm jump when \textit{inch-worm transformation} is activated. For navigation in \textit{mobile transformation}, we have proposed a path planning control framework. Visual inspection is performed using an Encoder-Decoder-based CNN \cite{Billah_ISVC2019}, which leads to highlighting defected regions (e.g. rust, crack, delamination) on steel surfaces. Moreover, a magnetic array based distance control is proposed in this work for autonomous magnetic adherence to steel surface. 

The rest of the paper is arranged as follows: Section \ref{S.2} discusses the overall proposed control architecture for the ARA robot. In section \ref{S.3}, the proposed control framework is described. The results of the different elements of the proposed control mechanisms are highlighted in section \ref{S.4}. Section \ref{S.5} concludes the primary findings and discuss some avenues for future research developments.

\section{Overall Architecture}\label{S.2}
\begin{figure}[ht]
\centerline{\includegraphics[width=1.0\linewidth]{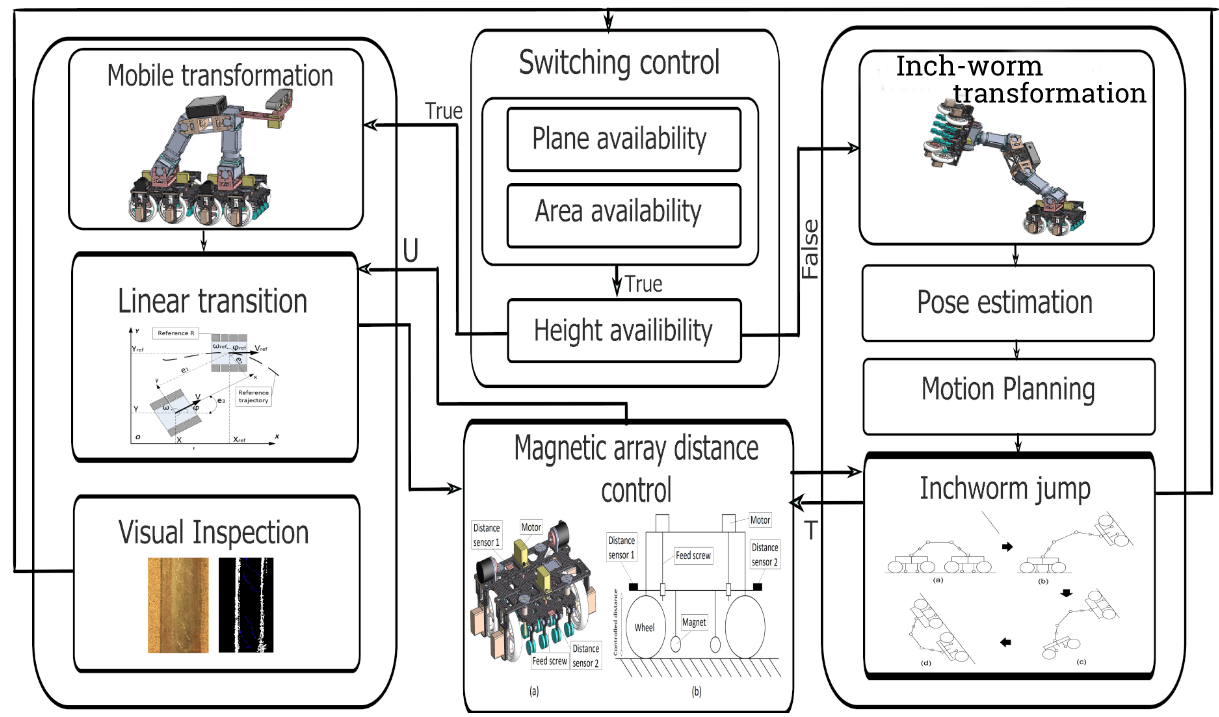}}
    \caption{The proposed control system framework for autonomous navigation}
    \label{fig:Overall_Framework}
\end{figure}

\begin{figure}[ht]
\centering
\setcounter{subfigure}{0}
\centerline{\includegraphics[width=0.74\linewidth]{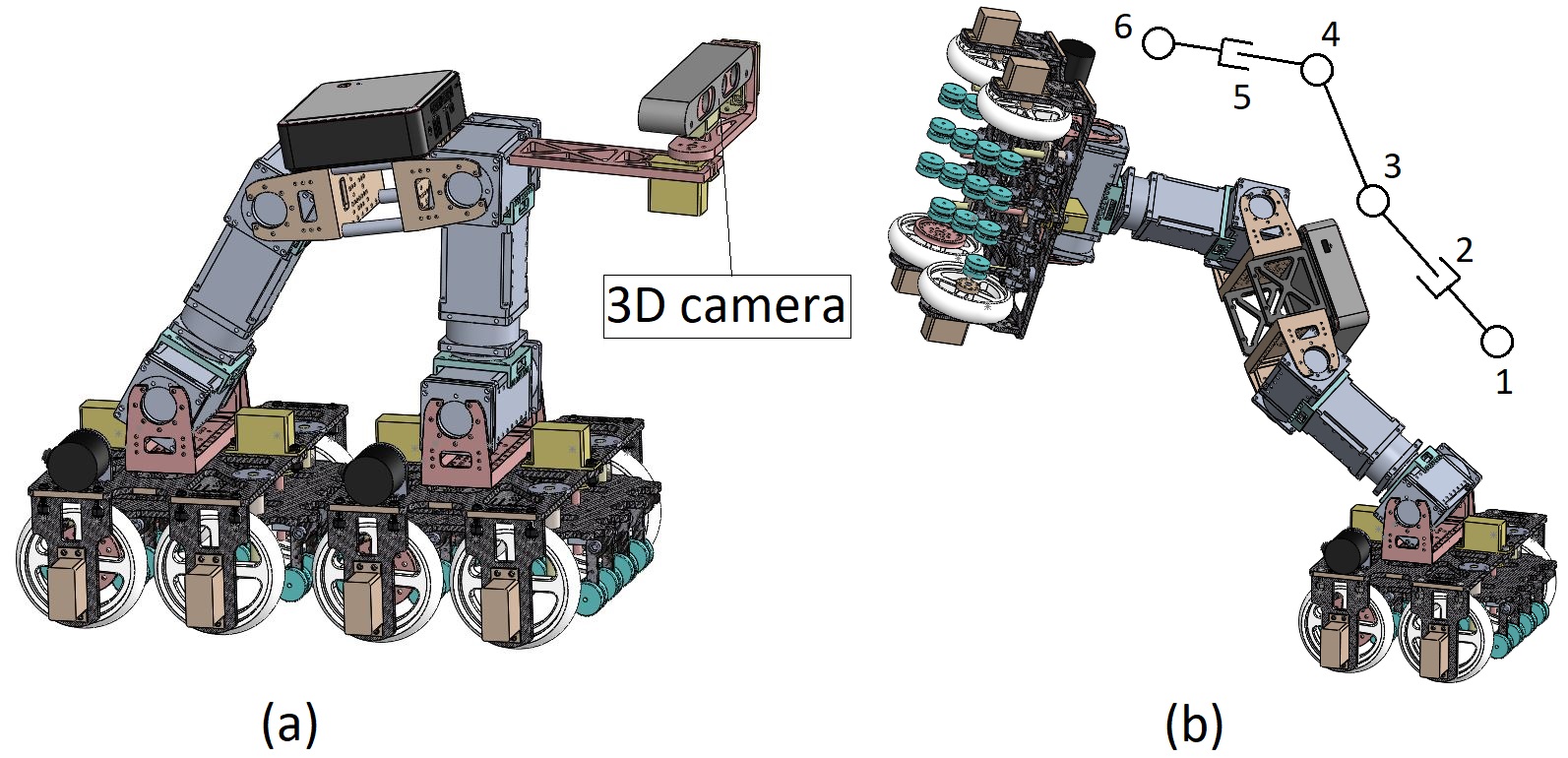}}
% \centerline{\includegraphics[width=1\linewidth]{images/mobilemode.JPG}}
\caption{ARA robot in (a) \textit{mobile} and (b) \textit{inch-worm transformation}}
\label{fig:newModule}
\end{figure}

ARA robot can configure itself into two transformations: mobile and inch-worm. In this work, we integrated a switching control mechanism (shown in Fig. \ref{fig:Overall_Framework}) to the robot. This control mechanism enables the robot to change its transformations depending on environmental conditions. When traversing on continuous and smooth steel surfaces, the robot activates the \textit{mobile transformation} as shown in Fig. \ref{fig:newModule}(a). The robot navigates using a path planning algorithm with the help of differential wheels and performs visual inspection of steel bridge structures. Moreover, the robot can move on an inclined steel surface by the adhesion forces supported by two magnetic arrays mounted on each robot foot. There are two working modes of the magnetic arrays: \textit{touched} and \textit{untouched}, indicating the distance from the magnetic arrays to the steel surface where the robot feet lies on. \textit{Touched mode} means the distance is zero, and the \textit{untouched} one keeps the distance around $1 mm$. The \textit{mobile transformation} requires both magnetic arrays operating in untouched mode to generate two magnetic adhesion forces, which are enough for the robot standing on the inclined surface, and in same time still let the robot can move by its wheels. The robot switches into \textit{inch-worm transformation} (Fig. \ref{fig:newModule}(b)) when it detects a complex steel surface, which cannot move on wheels, then activates an inch-worm jump to the next surface. As performing inch-worm jump, only one robot foot attaches on the steel surface. To create enough adhesion force for robot standing, the magnetic arrays is switched to \textit{touched mode}, which fully adheres the array to the steel surface. The switching control mechanism controls the movement of the robot, detects environment type and sends the appropriate command to executable nodes.

\begin{figure}[ht]
\centerline{\includegraphics[width=0.7\linewidth]{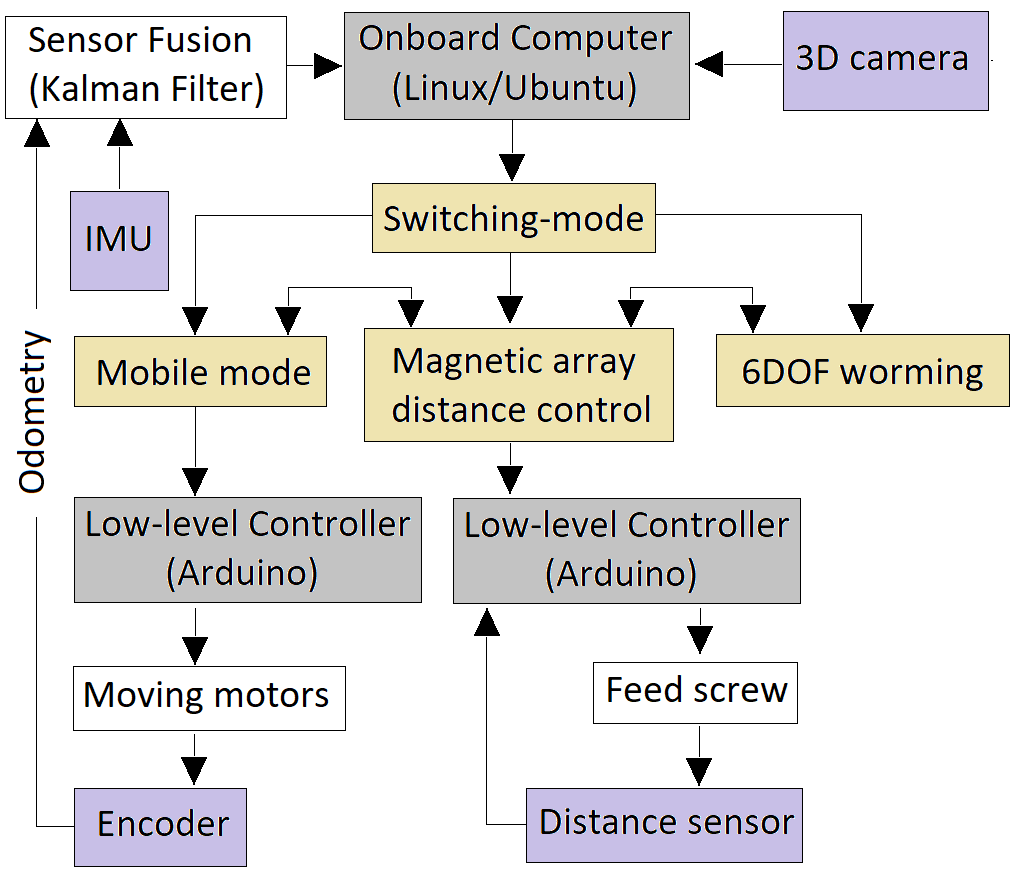}}
    \caption{The control architecture integrated into ARA robot \cite{icra20} }
    \label{fig:Control Structure}
\end{figure}

The control architecture of ARA robot is comprised of multiple low-level and high-level control structures. Several tasks are performed by the low-level control structure (Arduino). The wheel's velocity, encoder reading, Inertial Measurement Unit (IMU) sensor measurement  and the magnetic array function are performed in this control level. The high-level control embedded in an on-board processor manages switching control function, point cloud data processing, inverse kinematics and motion planning. The arrangement of the high level and low-level controls is shown in Fig. \ref{fig:Control Structure}. The control framework is described in detail in next section.

\section{Control System Framework}\label{S.3}
The ARA robot control framework composed of four modules: \textit{switching control}, \textit{magnetic array distance control}, \textit{mobile transformation}, and \textit{inch-worm transformation}. An overview of the overall framework is shown in Fig. \ref{fig:Overall_Framework}.

\subsection{Switching Control}
The switching control $S$ enables the robot to autonomously configure itself into two transformations (mobile and inch-worm). The control employs switching function $S$, represented in Eq.\ref{eq:switchingFunction}. The function takes as input three Boolean parameters: plane availability $S_{pa}$, area availability $S_{am}$ and height availability $S_{hc}$. These parameters determine if there is any still surface available, while enabling the estimation of the area of the surface and its height. A logical operation is performed on these parameters using function $f(.)$. The function parameters are estimated from 3D point cloud data of steel surface.
\begin{equation} \label{eq:switchingFunction}
S = f(S_{pa}, S_{am}, S_{hc}) = S_{pa} S_{am} S_{hc}.
\end{equation}
The robot configures to \textit{mobile transformation} if the function return a true value. The false value configures the robot into the \textit{inch-worm transformation}.

\textbf{Plane availability:} The 3D point cloud of steel surface is processed using \textit{pass-through} filtering, \textit{downsampling}, and \textit{plane detection} \cite{buiIRC2020Sort}. The \textit{plane detection} applied the RANSAC method extracts the planar point cloud $P_{cl}$ from the initial point cloud. The plane availability is checked using  Eq. (\ref{Eq:PlaneCheck}):
\begin{equation}\label{Eq:PlaneCheck}
\begin{cases}
    S_{pa} = False, \, if \, P_{cl} = \emptyset \\
    S_{pa} = True, \, otherwise.
\end{cases}
\end{equation}
Moreover, two functions \textit{get\_centroid} and \textit{get\_normal\_vector} provides the point cloud's centroid $C_{P_{cl}}$ and normal vector $\vec{N_{P_{cl}}}$ of the point cloud.

\textbf{Area availability:} The robot feet requires an area estimation of the available planar surface area $P_{cl}$. It is essential to ensure the availability of sufficient area for successful robot transition. This is a popular problem in legged-robot, which is investigated carefully in \cite{deits2015computing, jatsun2017footstep, hildebrandt2019versatile}. In \cite{deits2015computing, jatsun2017footstep}, the authors proposed the convex-based algorithms, which deployed convex optimization problem to determine an obstacle-free ellipsoid (convex one), then estimate step-able areas for biped robot. In \cite{hildebrandt2019versatile}, the authors proposed an algorithm to determine the valid convex collision-free regions with geometrical constraints of obstacles. In those algorithms, a portion of the step-able area, especially as the vertex number is small, was not considered due to the convex approximation. It is a problem for our inspection robot with large feet pair due to limited step-able areas on steel bridges. These algorithms are not possible to find a step-able area for the robot to worm in many cases.
Thus, we developed two algorithms, which can efficiently estimate a step-able area for the inspection robot while the input planes are non-convex. The first algorithm - \textit{Algorithm \ref{alg:boundaryestimation}} extracts a non-convex boundary from the planar point cloud, then the second one - \textit{Algorithm \ref{alg:areaestimation}} checks the sufficiency of the available planar surface. 

\begin{figure}[ht]
\centerline{\includegraphics[width=0.7\linewidth]{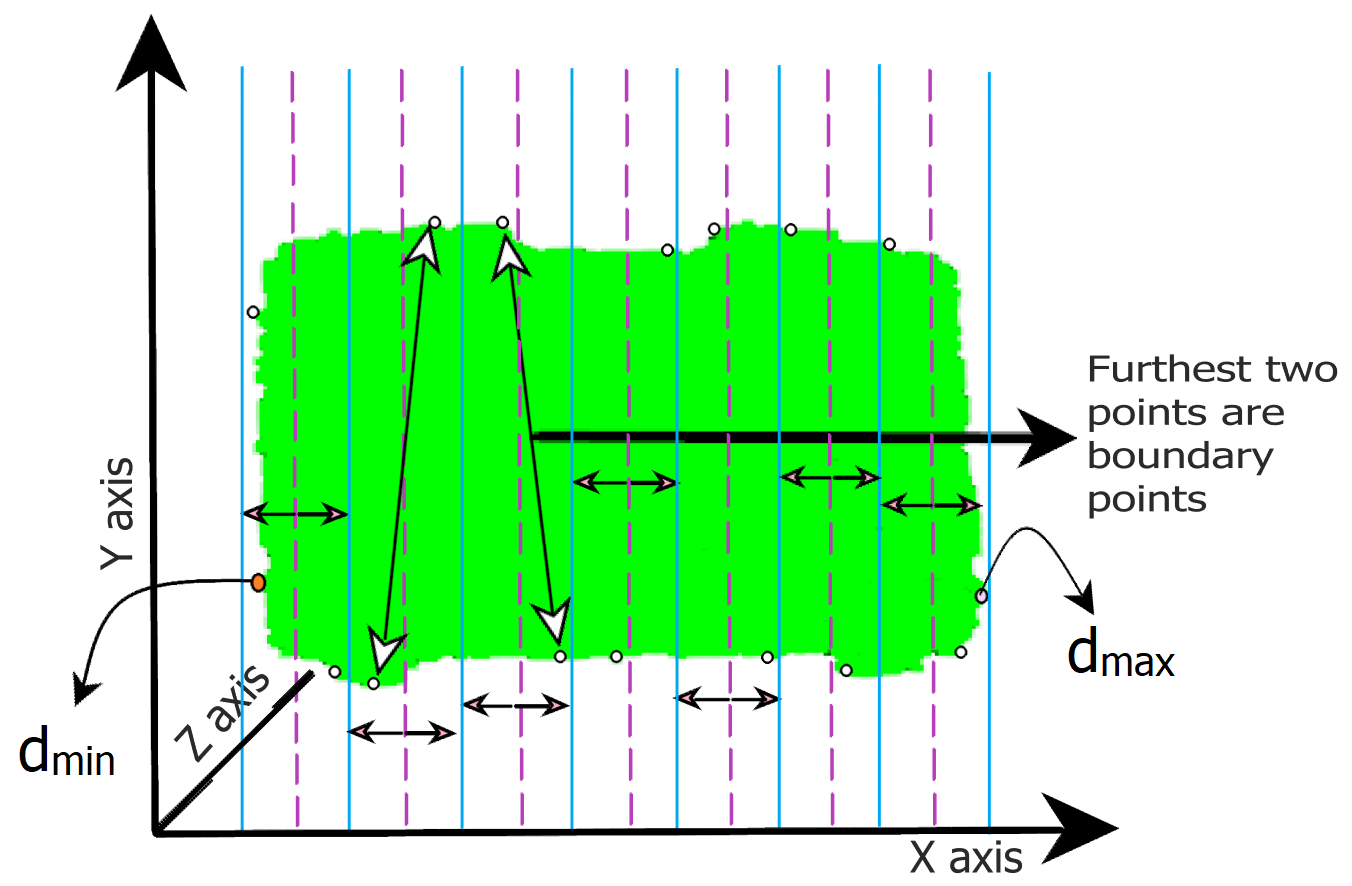}}
    \caption{ Boundary point estimation from 3D point cloud data}
    \label{fig:BoundaryAlgorithm}
\end{figure}

\begin{algorithm}
\small
\caption{Non-convex boundary point estimation from 3D point cloud data of steel bridges}\label{alg:boundaryestimation}
    \begin{algorithmic}[1]
        \Procedure{BoundaryEstimation}{$P_{cl}, \alpha_s$}
            \State $Planes$ = \{$xy, yz, zx$\}
            \State $d_{min}$ = $\forall_{i \in Planes}$ {\ssmall //Point along minimum value of plane $i$}
            \State $d_{max}$ = $\forall_{i \in Planes}$ {\ssmall //Point along maximum value of plane $i$}
            \State Initialize $B_s = \{\}$
            \For{$p \in Planes $}
                \State $i \to 1$
                \While {$sl_{p_i} < d_{max}$}
                    \State $sl_{p_i} = d_{min_p} + i*\alpha_s$
                    \State $PS_{p_i}$ = Set of points in range  $sl_{p_i}\pm \alpha_s/2$
                    % \State   \begin{equation} \begin{split}                  \vspace{0.5cm}P_{cl_{A}}, P_{cl_{B}} = 
                    %     \underset{P_i,P_j}{\mathrm{argmax}} \{d = \left\Vert{P_i-P_j}\right\Vert | P_i \in \\ PS_{p_i}, P_j \in PS_{p_i}\}\end{split}\end{equation} 
                \State { $P_{cl_{A}}, P_{cl_{B}} = \underset{\forall\{P_i,P_j\} \in PS_{p_i}}{\mathrm{argmax}} \{ \left\Vert{P_i-P_j}\right\Vert$}\}
                    \State $B_s = B_s \cup \{P_{cl_{A}}, P_{cl_{B}}\}$ 
                    \State $i = i+ 1 $
                \EndWhile
            \EndFor \\
            \Return $B_s$
        \EndProcedure
    \end{algorithmic}
\end{algorithm}
The boundary points estimated in \textit{Algorithm \ref{alg:boundaryestimation}} are performed by a window-based approach. The algorithm's input is the point cloud $P_{cl}$ of the estimated planar surface and a slicing parameter $\alpha_s$. At first, we calculate the two furthest points represented as $d_{min}$ and $d_{max}$ in the point cloud $P_{cl}$ along each plane. Then the point cloud is divided into multiple smaller slices along the three planes. For each slice in a particular plane $p$ we calculate the slicing index  $sl_{p}$, which represents the center coordinate of the slice as shown in line $9$ of algorithm \ref{alg:boundaryestimation}. After that, the point sets $PS_{p}$ in the range $sl_{p}\pm \alpha_s/2$ is extracted from $P_{cl}$. This sliding factor is experimentally determined based on the point cloud size. 
For each set of points from $PS_{p}$, we extract the two furthest points ($P_{cl_{A}}, P_{cl_{B}}$). These two points are estimated as the boundary point for that particular slice and added to the boundary point sets $B_s$. This approach helps the algorithm work well with plane with small holes inside. A pictorial representation of the boundary estimation algorithm is shown in Fig. \ref{fig:BoundaryAlgorithm}.

\begin{algorithm}
\small
\caption{Area Checking from the plane surface boundary points and Pose Calculation}\label{alg:areaestimation}
    \begin{algorithmic}[1]
        \Procedure{Area}{$B_{s}, C_{P_{cl}}, \vec{n_{P_{cl}}}, w, l, t, S_{am}$}
            \State $N_{clos}$ = Find $n$ closest points to $C_{P_{cl}}$ from $B_{s}$
            \For { $N_i \in N_{clos}$ }
                \State $R = \{\}$, {\ssmall //Estimated rectangle corner points}
                \State $\vec{e_{x_i}} = N_i - C_{P_{cl}} $
                \State $\vec{e_{z_i}} = \vec{n_{P_{cl}}}$
                \State $\vec{e_{y_i}} = \vec{e_{x_i}} \times \vec{e_{y_i}} $
                % \State $e = e \cup \{\vec{e_{x_i}}, \vec{e_{y_i}}, \vec{e_{z_i}}\}$
                \State $k_w = \frac{w}{|\vec{e_{x_i}}|} e_{y_i} $ and $k_l = \frac{b}{|\vec{e_{y_i}}|} e_{x_i}$,
                \State $\{R_1, R_2\} = \{ N_i + k_w, N_i - k_w \}$
                \State $R = R \cup \{R_1, R_2\} $
                \State $R = R \cup \{R_1+k_l, R_2+k_l\}$
                \State $M = \forall_{r_i \in R} \{ \frac{r_i + r_{i+1}}{2}\}$
                \State $R = R \cup M$
                \State $S_{am} = True$
                \For { $r_i \in R$ }
                    \State $Q_i$ = Find $m$ closest points to $r_i$ 
                    \State $d_{r_i} = \left\Vert{d_{r_i}, C_{P_{cl}}}\right\Vert$ and $d_{Q_i} = \left\Vert{Q_i, C_{P_{cl}}}\right\Vert$
                    \State $S_i = (d_{r_i} < d_{Q_i}) \lor(\frac{d_{r_i}-d_{Q_i}}{{d_{r_{i}}}} <t$
                    % \State $S_{am_i} = (d_{r_i}<d_Q_i)\lor(\frac{d_{r_i}-d_{Q_i}}{{d_{r_{i}}}} <t)$
                    \State $S_{am} = S_{am} \land S_i$ 
                \EndFor
                \State Pose = \{Orientation, Position\}
                \If{ $S_{am} == $ True}
                    \State $R_c$ = Centroid of $R$
                    \State Orientation = ($\vec{e_{x_i}}, \vec{e_{y_i}},  \vec{e_{z_i}}$)
                    \State Position = $(x_{R_c}, y_{R_c}-l/4, z_{R_c})$
                    \State \Return Pose
                \EndIf
            \EndFor
            \State \Return False 
        \EndProcedure
    \end{algorithmic}
\end{algorithm}

After estimating the boundary points $B_s$, we estimate the area availability parameter $S_{am}$. The estimation is performed using the \textit{Algorithm \ref{alg:areaestimation}}. The input of this algorithm are the boundary points $B_s$, point cloud centroid $C_{P_{cl}}$, normal vector of point cloud $\vec{n_{P_{cl}}}$, length $l$ and width $w$ of robot leg and wheel distance tolerance $t$. At first we calculate the $n$ closest points ($N_{clos}$) from $B_s$ to  the point cloud centroid $C_{P_{cl}}$.  For each point, $N_i$ in the set $N_{clos}$, a set of computations is performed to estimate the plane corners for adherence to robot wheels. At first, the coordinate frame vectors $\vec{e_{x_i}}, \vec{e_{y_i}}$ and $\vec{e_{z_i}}$ are calculated for point $N_i$. In the next step, the algorithm estimates the corner points of a rectangle of width $w$ and length $l$, which is also robot foot width and length, respectively. We estimate the rectangle edges parallel along the vectors $\vec{e_{x_i}}$ and $\vec{e_{y_i}}$. Therefore, the four corners $R$ of the rectangle are estimated using these two vectors. Additionally, we include four middle points of the estimated rectangle corners in $R$ to alleviate point cloud collection error as well as accommodate for the non-convex shape of the steel surface. After the estimation step, we find $m$ closest points to $R$ from the $B_s$ to measure if the points in $R$ are inside the boundary. Hence, we calculate the distance from point cloud centroid $C_{P_{cl}}$ to $R$ and $Q$, respectively. The algorithm considers a point lies inside the boundary if its tolerance is less than $t$ and the distance to centroid should be less than its neighbors. The algorithm's performance is presented in Fig. \ref{fig:boundaryRectangle}(b)-(d). When the value of $S_{am}$ is true for all the conditions, we consider those sets of points as rectangular points.

\textbf{Height availability:} The height availability $S_{hc}$ is crucial for the switching control. Based on this parameter, the switching control activates the robot transformations. At first the point cloud's centroid $C_{P_{cl}}$ is calculated along the camera frame $f_c$. Then it is transformed to the robot base frame $f_{rb}$ using Eq. \ref{Eq:transMatrix}.
\begin{equation}\label{Eq:transMatrix}
    P_{C_{f_{rb}}} = T_{f_{rb}f_c} P_{C_{f_c}},
\end{equation}
where $(p_{C_{f_{rb}}}$, $p_{C_{f_c}})$ is coordinate of the centroid $C_c$ in the camera frame and the robot base frame, respectively.  $T_{f_{rb}f_c}$ is the transformation matrix from the camera frame $f_c$ to the robot base frame $f_{rb}$.

The plane height $z_{f_{rb}}$ coordinate is then compared with the robot base height. If they are equal, the returned result is \textit{true} and the robot is configured as \textit{mobile transformation}. Otherwise, it returns \textit{false} and the robot go to \textit{\textit{inch-worm transformation}}. The height availability condition is shown as in Eq. \ref{Eq:heightCheck}.
\begin{equation}\label{Eq:heightCheck}
\begin{cases}
    S_{hc} = True, \, if \,{z_{f_{rb}}} = z_{robotbase}\\
    S_{hc} = False, \, otherwise.
\end{cases}
\end{equation}

\subsection{Magnetic Array Distance Control}
\begin{figure}[ht]
\centering
\setcounter{subfigure}{0}
\centerline{\includegraphics[width=0.9\linewidth]{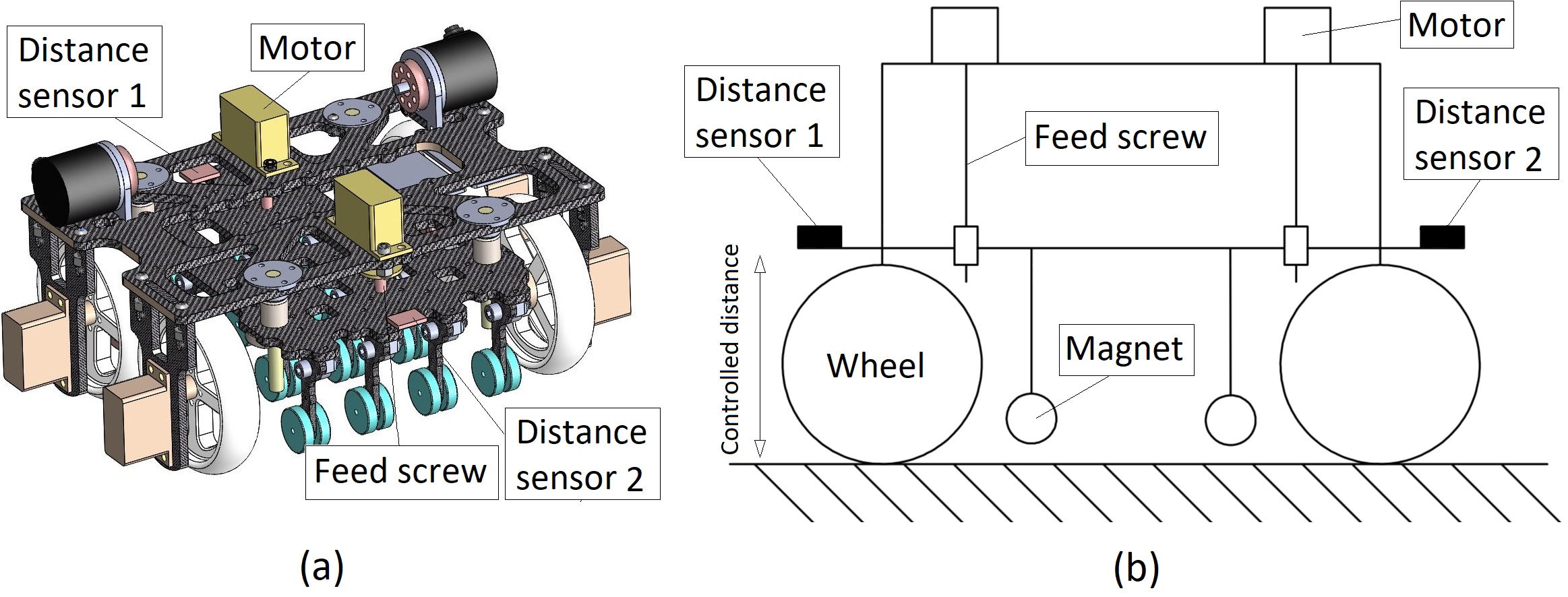}}
\caption{Magnetic array distance control system: a) 3D model b) 2D diagram}
\label{fig:distancecontrol}
\end{figure}

The magnetic arrays on two feet of ARA robot provides two working modes: touched and untouched. A PID distance controller shown in Fig.\ref{fig:distancecontrol} manipulates the magnetic arrays to move up and down to adjust the adhesion forces. The distances are sensed by two distance sensors mounted on two sides of each magnetic array, which are actuated by a pair of DC-motors and parallel screws. The distance of $1mm$ in \textit{untouched mode} is critical for robot to traverse on a slope steel surface, thus the controller's accuracy is crucial. A minor mechanical error may lead to part of the magnetic array transform into the touching mode and result in a huge load to the wheels' motors. Therefore, the symmetrical design of the distance sensors improves the accuracy of the controller significantly to avoid the aforementioned difficulties. 

\subsection{Mobile transformation}
\begin{figure}[ht]
\centerline{\includegraphics[width=0.55\linewidth]{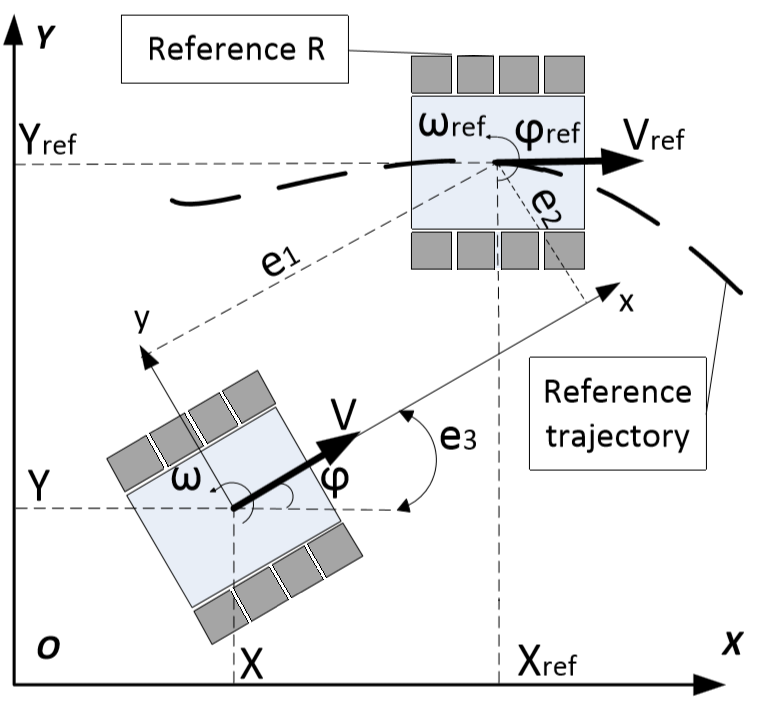}}
    \caption{Statement of the problem of robot trajectory movement control}
    \label{fig:path}
\end{figure}

ARA robot switches to \textit{mobile transformation} to move on the continuous steel surface. The transition in the smooth surface is performed by the two-wheel of each robot foot. To perform the navigation smoothly, we defined the problem statement of two-wheel movement as in Fig. \ref{fig:path}. 

The dynamic equation of the robot is derived from \cite{Guo_ICRA14}. The current position and the orientation of the robot is represented with (${x}_{c}$, ${y}_{c}$, ${\varphi}_{c}$). The robot needs to move to a target position $R(x_r,y_r)$ with constant velocity $v_r$ with the heading angle $\phi_r$.

\begin{figure}[ht]
\centerline{\includegraphics[width=0.9\linewidth]{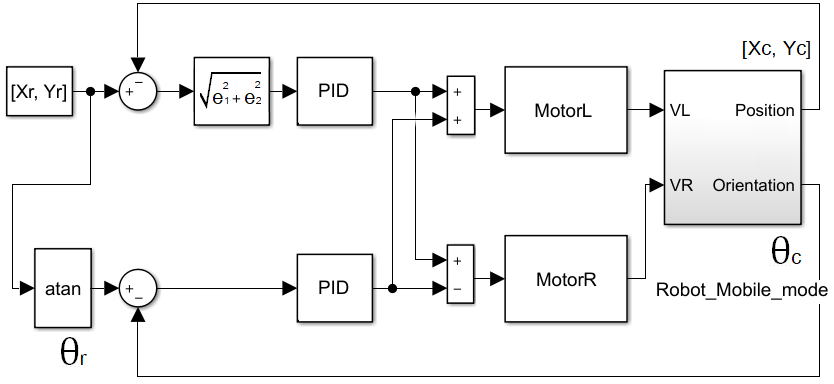}}
    \caption{The mixed PID controller for path planning in mobile mode}
    \label{fig:mobilemodePID}
\end{figure}

To navigate the robot to point $R$, a control is designed to help the robot to track the point $R$. A tracking error vector is introduced $e = [e_1,e_2,e_3]^T $, presented as in Eq. (\ref{matrixd}).

\begin{equation}
\begin{bmatrix} \label{matrixd}
    e_1     \\
    e_2    \\
    e_3
 \end{bmatrix}
=
\begin{bmatrix}
   \cos\varphi_{c} & \sin\varphi_{c} & 0    \\
   -\sin\varphi{c} &  \cos\varphi_{c} & 0  \\
    0 & 0 & 1
 \end{bmatrix}
\begin{bmatrix}
x_r - x_c\\
y_r - y_c\\
\varphi_r -\varphi_c
\end{bmatrix}
\end{equation}
Derivative of Eq. (\ref{matrixd}) and linearizing \cite{Guo_ICRA14}, we get Eq. \ref{matrixe}:

% \begin{equation}
% \begin{bmatrix} \label{Eq:matrixd}
%     \dot{e}_1     \\
%     \dot{e}_2     \\
%     \dot{e}_3
%  \end{bmatrix}
% =
% \begin{bmatrix}
%   e_{2}\omega_c - v_c + v_r\cos{e}_{3}   \\
%   -e_1\omega_{c} +  v_r\sin e_{3}  \\
%     \omega_r - \omega_c
%  \end{bmatrix}
% \end{equation}

\begin{equation}
\begin{bmatrix} \label{matrixe}
    \dot{e}_1     \\
    \dot{e}_2     \\
    \dot{e}_3
 \end{bmatrix}
=
\begin{bmatrix}
  -1  \\
    0  \\
    0
 \end{bmatrix}
 v_c
 +
\begin{bmatrix}
e_2 \\
-e_1\\
-1
\end{bmatrix}
%
% \omega_c
+
\begin{bmatrix}
    v_r\cos{e}_1     \\
    v_r\sin{e}_2     \\
    \omega_r
 \end{bmatrix}
\end{equation}
To track the reference point $R$ from the current position $C$, the error vector need to go to 0: $e_i \rightarrow 0$. A mixed PID controller designed for this purpose was shown in Fig.\ref{fig:mobilemodePID}. The trajectory is divided into discrete tracking positions as a set point of the first loop controller. The second loop handles the robot heading error obtained by $atan(y_r-y_c,x_r-x_c)$. A mixer combines the two controllers' outputs, then feed the velocities to the motors. The robot's position and orientation feedback is read by the encoders and IMU sensors. 

\begin{figure}[ht]
\centerline{\includegraphics[width=0.9\linewidth, height = 0.4\linewidth]{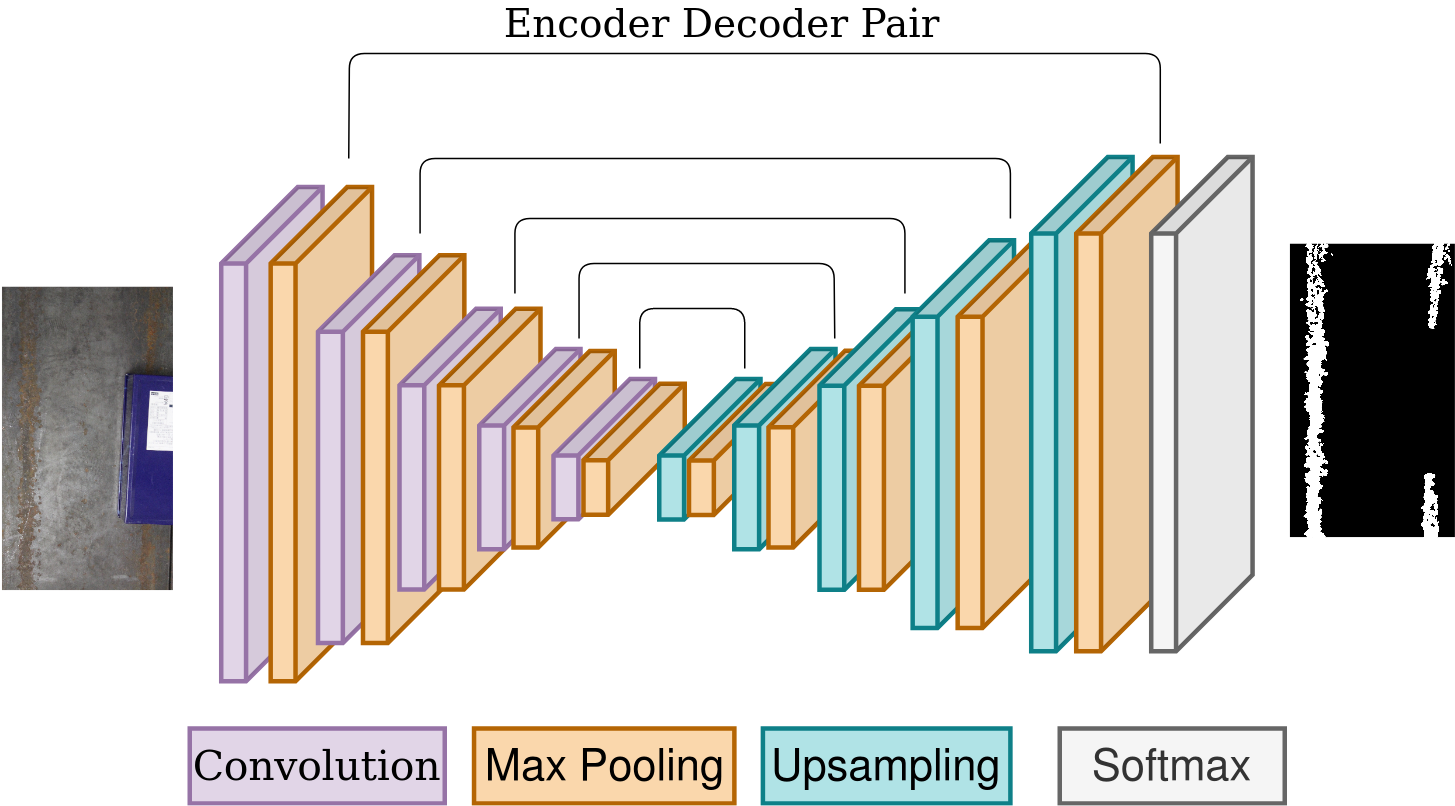}}
    \caption{An encoder decoder based CNN architecture for steel bridge inspection}
    \label{fig:inspectionFramework}
\end{figure}

\textbf{The inspection framework:} 
After performing the linear transition, the robot conducts the visual inspection. The RGB camera captures images from the steel surface and sends them to an encoder-decoder based CNN \cite{Billah_ISVC2019}. The CNN architecture segments images into defect and healthy regions. The input images are passed through five encoder layers followed by five decoder layers. 
Each encoder performs a $7 \times 7$ convolution operation, to extract defect feature maps. The convolution operation is followed by a batch normalization operation and ReLU activation. The feature space is down-sampled using a $2 \times 2$ max-pooling unit and fed to the next encoder layer. Five encoder layers are utilized for feature extraction. Once the last encoder layer is reached, the resultant feature maps enter the decoder portion of the network, where they are up-sampled using bi-linear interpolation in each decoder layer. The up-sampling operations are again followed by a convolution operation, batch normalization, and ReLU activation. A graphical representation of the CNN architecture is shown in Fig. \ref{fig:inspectionFramework}. 
The CNN architecture is pre-trained on $3000$ steel images containing severe defects (e.g. corrosion). The training and data preparation is extensively elaborated in \cite{Billah_ISVC2019}. For the hyperparameter optimization, we used the ADAM optimizer with a learning rate of $0.0001$. The network was trained for $150$ epochs in this experiment. 

\subsection{Inch-worm transformation}\label{worming mode}
\begin{figure}[ht]
\centerline{\includegraphics[width=0.7\linewidth, height=0.55\linewidth]{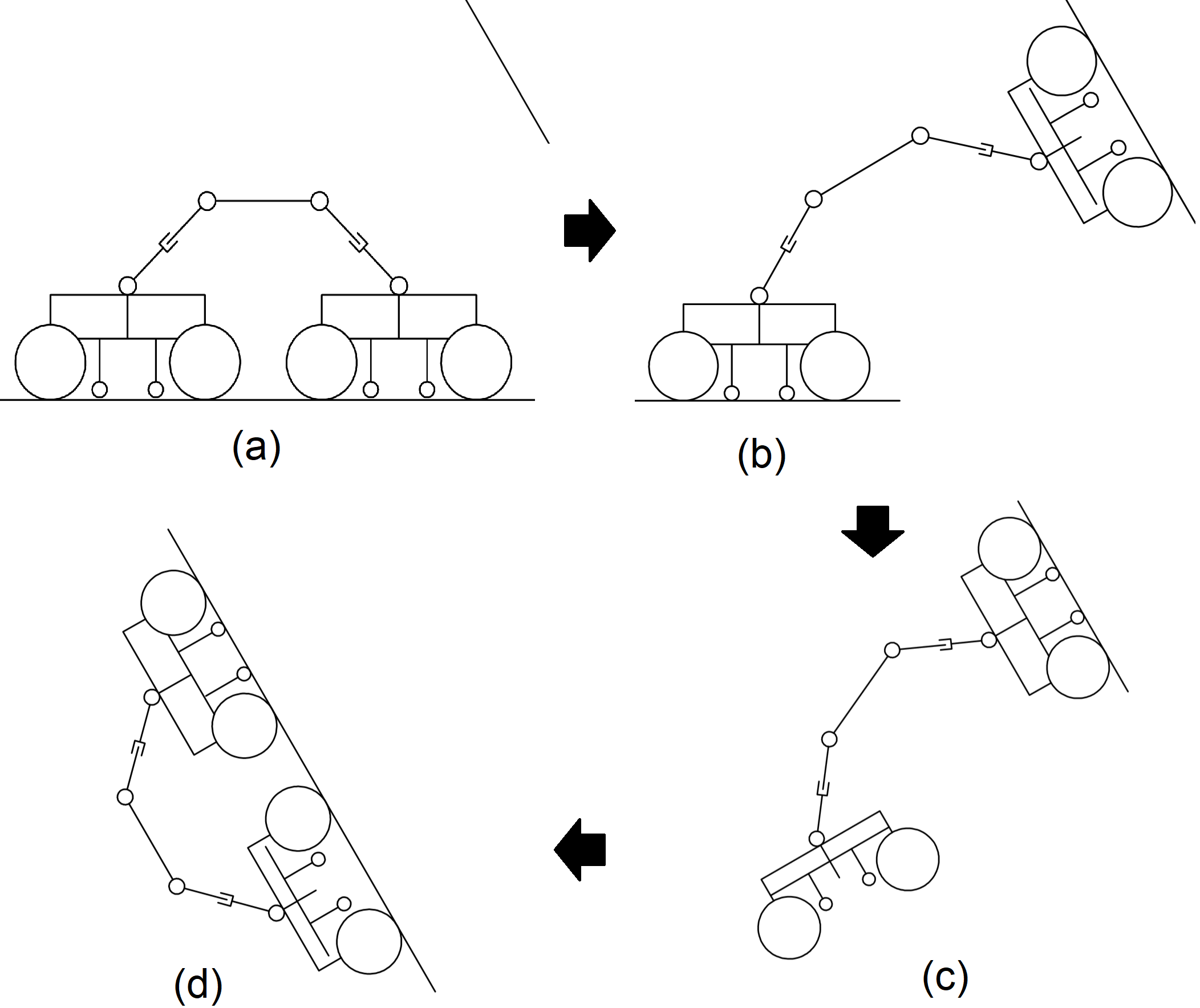}}
\caption{Inch-worm jump from one steel surface to another}
\label{fig:inchWormJump}
\end{figure}

The inch-worm transformation enables the robot to perform an inch-worm jump from one steel surface to another as shown in Fig. \ref{fig:inchWormJump}. At first, the permanent magnet on the second robot foot is set to \textit{touched mode}, which adheres the leg on steel surface and generates a strong adhesive force for the robot to stand and perform the worming. A controller manipulates the joints to move the first robot leg towards the target plane as shown in  Fig. \ref{fig:inchWormJump}(b). As the first leg touches the target surface, the first and second permanent magnets are switched to \textit{touched mode} and \textit{untouched mode}, respectively. After that, the second leg detaches from the starting surface as in Fig. \ref{fig:inchWormJump}(c). Finally, in Fig. \ref{fig:inchWormJump}(d) the second leg is adhered the target surface.

Converting from \textit{mobile} (Fig. \ref{fig:newModule}(a)) to \textit{inch-worm transformation} is challenging for the motion planner to create a trajectory. To have a better performance, a convenient robot pose $P_{conv}$ is proposed where the robot should move to firstly as starting of \textit{inch-worm transformation}. From there, the motion planner will generate a trajectory for the first leg to move the destination. The worming is completed by moving the second leg to the target surface and reform \textit{mobile configuration}. 
To have the target pose, the robot needs to determine the target plane and its pose, which are an output of \textit{Algorithm \ref{alg:areaestimation}}.

The flexibility of the robot in worming is made of the six DoF arm. The revolute joints of the arm in Fig. \ref{fig:newModule}(b) can rotate along three different axes separately. For example, the joint $2$ and joint $5$ in Fig. \ref{fig:newModule}(b) are configured to rotate around \textit{y-axis}. The rest of the joints are positioned to rotate around \textit{z-axis}. This configuration was selected for maintaining symmetry so that the manipulator can move efficiently in both worming and mobile mode. Our previous research states an in detail elaboration of the robotic arm in \cite{icra20}. 

\section{Experiment Results}\label{S.4}
In this paper, the experiment is implement on ARA robot version 1.0, which is derived from \cite{icra20} with an additional camera module. An RGB-D camera (ASUS Xtion Pro Live) is attached to the robot for point cloud collection and visual inspection. The camera calibration's parameters are integrated from the method implemented by \cite{buiIRC2020Sort}. The robot is localized by \textit{aruco marker}, which is placed on the robot standing surface. We performed a geometric calculation to locate the position between aruco marker and the robot base. Additionally, an Intel NUC 5 - Core i5 vPro was incorporated for employing the robotic operating system (ROS) as well as the inspection framework. 

We perform our experiment on two individual steel slabs located perpendicularly from each other. The steel slabs are highly corroded to replicate steel defects. The following section describe the experiment and their results elaborately.

\subsection{Switching control}
\begin{figure}[ht]
\centering
\includegraphics[height=0.42\linewidth]{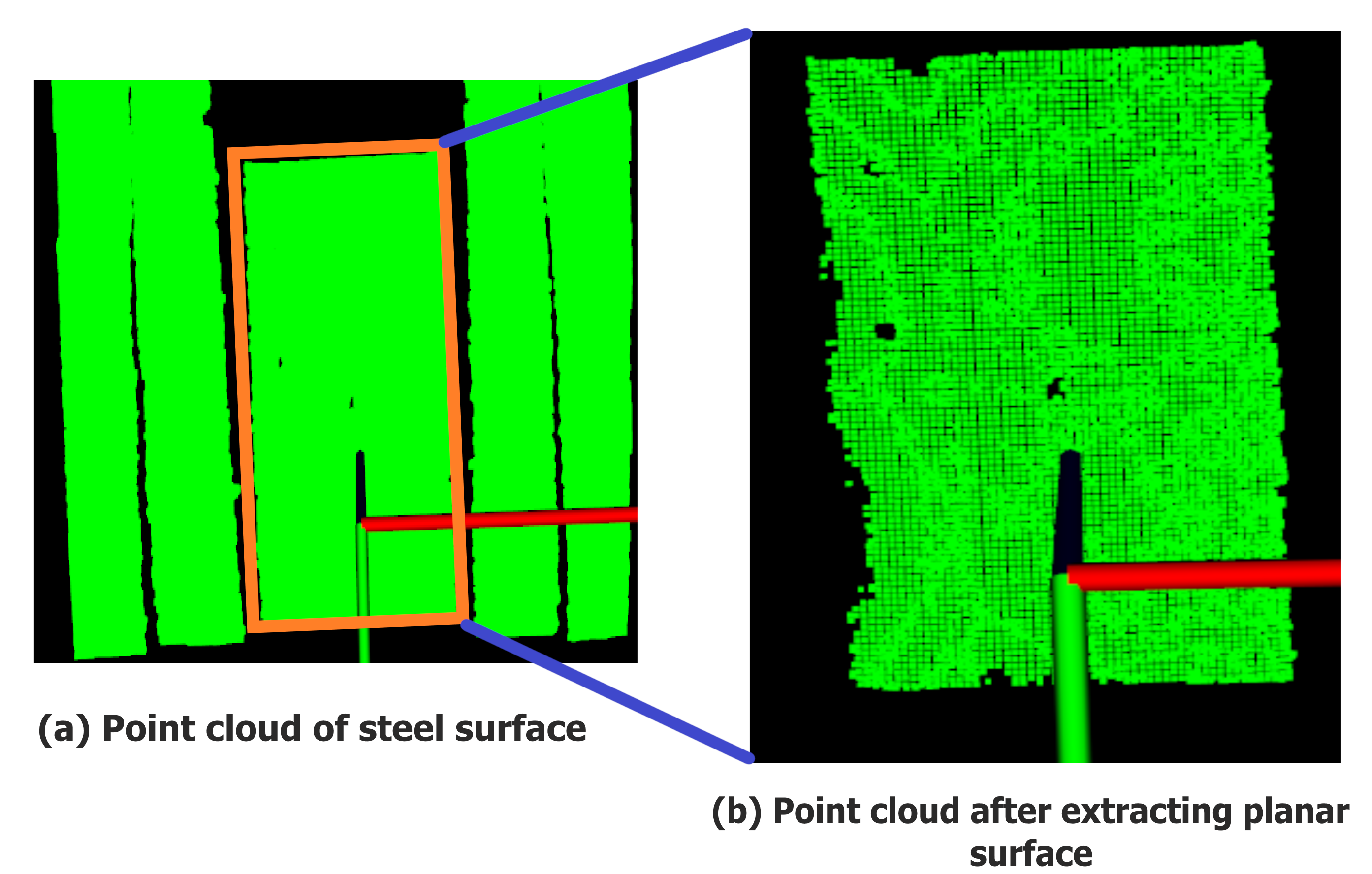}
\caption{Planar surface extraction from 3D point cloud of steel surface}
\label{fig:initialPointCloud}
\end{figure}

\begin{figure}[ht]
\centering
\setcounter{subfigure}{0}
\subfigure[]{\includegraphics[width=0.22\linewidth, height =0.35\linewidth]{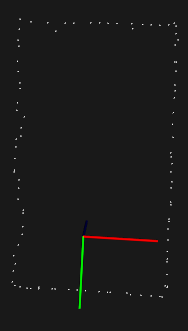}}
\subfigure[]{\includegraphics[width=0.24\linewidth, height =0.35\linewidth]{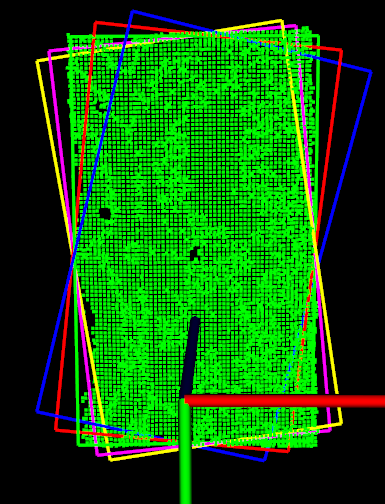}}
\subfigure[]{\includegraphics[width=0.22\linewidth, height =0.35\linewidth]{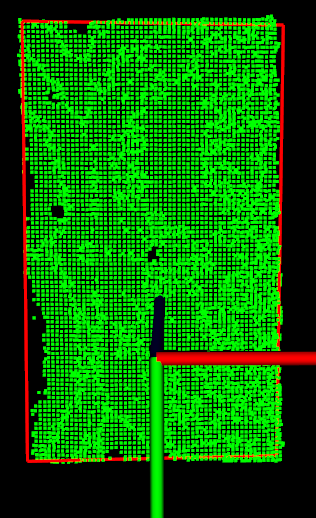}}
\subfigure[]{\includegraphics[width=0.26\linewidth, height =0.35\linewidth]{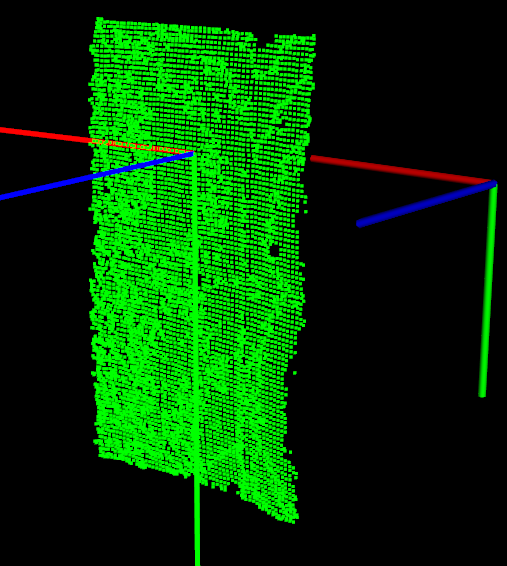}}
\caption{(a) Boundary set, (b) Area rectangle set, (c) The selected area rectangle, and (d) Pose estimation}
\label{fig:boundaryRectangle}
\end{figure}

At starting, the RGB-D point cloud data of a bridge steel bar was collected from the robot camera. An example of the initial point cloud is shown in Fig.\ref{fig:initialPointCloud}(a). After performing some pre-processing operations such as \textit{pass-through filtering} and \textit{downsampling}, the data is sent to \textit{plane detection} to extract the planar surface. The processed point cloud is shown in Fig.\ref{fig:initialPointCloud}(b). The coordinate frame is also shown in the figure with \textit{x-}axis in red, \textit{y-}axis in green, and  \textit{z-}axis in blue.
After obtaining the planar surfaces,  the surface boundary points and \textit{Area availability} checking were performed by \textit{Algorithm \ref{alg:boundaryestimation}} and \textit{Algorithm \ref{alg:areaestimation}} on two different surfaces, one containing sufficient area for movement and the other without. Using \textit{Algorithm \ref{alg:boundaryestimation}}, two boundary points of the two different point cloud as shown in Fig.\ref{fig:boundaryRectangle}(a) and Fig.\ref{fig:boundaryRectangle}(c). 
The area availability check from \textit{Algorithm \ref{alg:areaestimation}} is employed using the boundary point estimated. The algorithm parameters were as following : $n = 5, m = 3 $ and $t = 0.02$. Five rectangles were estimated for the robot feet with this algorithm shown in Fig. \ref{fig:boundaryRectangle}(b)  in red, yellow, blue, green, and purple color. In Fig. \ref{fig:boundaryRectangle}(b), several corners of all the red, yellow, purple, and blue rectangles were outside of the point cloud area. It represented that these rectangles area were not sufficient enough for an inch-worm jump. Only the green rectangle was inside the point cloud, satisfying area requirement. The selected rectangle is shown in Fig.  \ref{fig:boundaryRectangle}(c) (in red color). Since there is enough area for an inch-worm jump the variable $S_{am}$ is set to true by the algorithm. After that, the planar surface pose was estimated as shown in Fig. \ref{fig:boundaryRectangle}(d) with three orientations shown in red, green, and blue color on the point cloud surface.
 
\begin{figure}[ht]
\centering
\setcounter{subfigure}{0}
\subfigure[]{\includegraphics[width=0.4\linewidth, height = 0.23\linewidth]{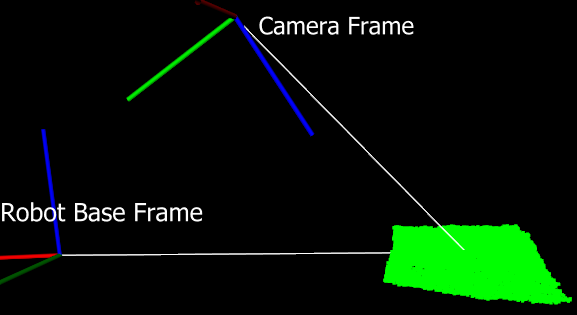}}
\subfigure[]{\includegraphics[width=0.38\linewidth, height = 0.23\linewidth]{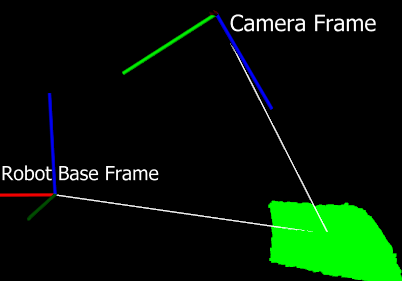}}
\caption{Surface Height Check (a) Same Height  \& (b) Different Height}
\label{fig:checkHeight}
\end{figure}

The surface pose was then transformed into the robot base frame. If the pose's height (corresponding to \textit{z-} axis in the robot base frame) was equal to robot base height, the value of $S_{hc} $ was set to \textit{True} and the robot configured itself as \textit{mobile transformation}. Fig. \ref{fig:checkHeight}(b) represented another scenario as the point cloud is from a surface, which was $d=7cm$ lower than the robot base. In this case, the heights were different, then the returned value of variable $S_{hc}$ was \textit{false}, and robot performs \textit{inch-worm transformation} in the next step.

\subsection{Navigation in mobile and \textit{inch-worm transformation}}
\begin{figure}[ht]
\centerline{\includegraphics[width=0.85\linewidth]{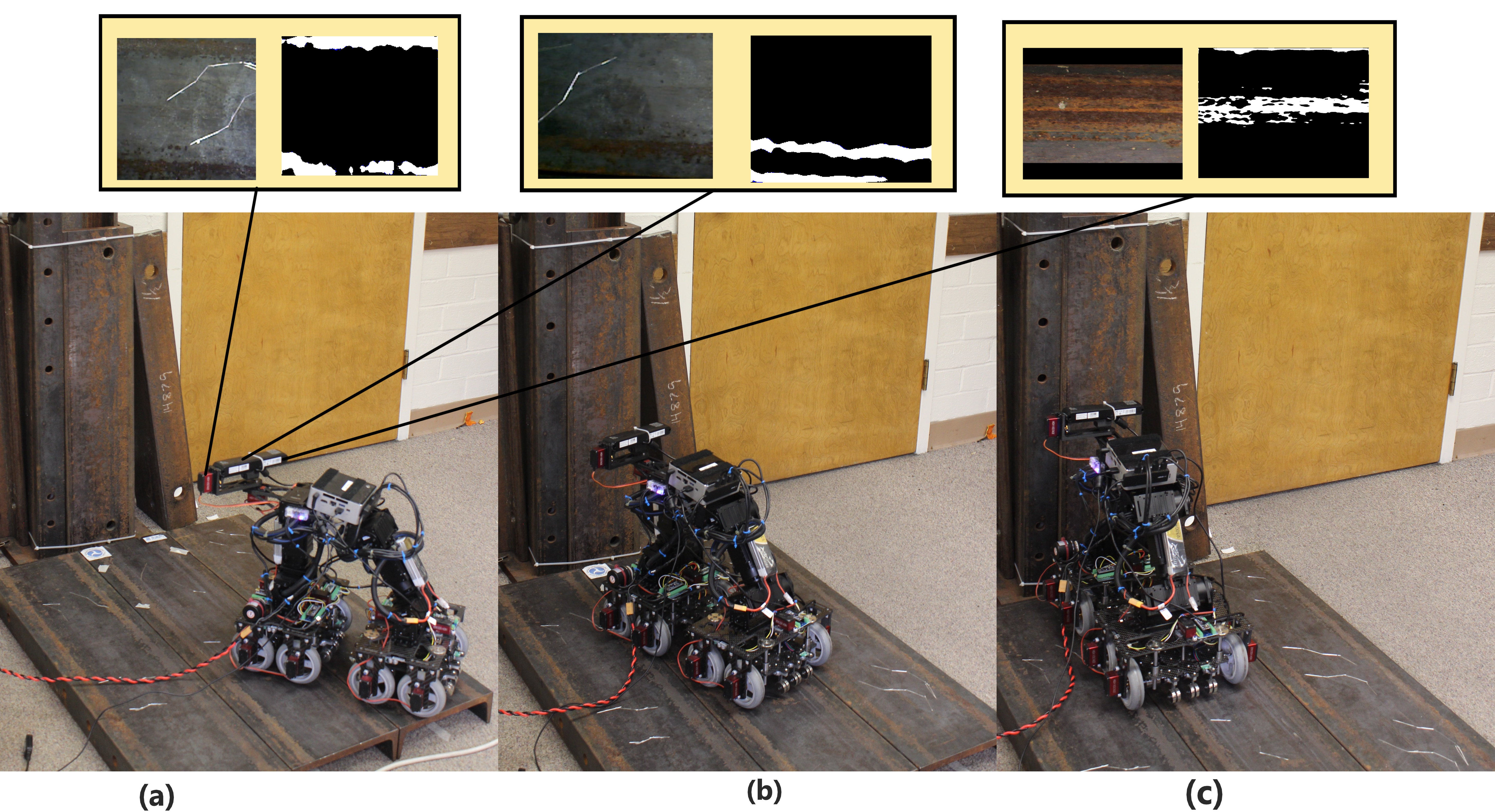}}
    \caption{Robot movement in \textit{mobile transformation} and visual inspection}
    \label{fig:mobileMove}
\end{figure}

In \textit{mobile transformation} the robot can move both in \textit{x-} and \textit{y-} directions simultaneously. The navigation in \textit{mobile transformation} is represented in Fig. \ref{fig:mobileMove}. In this transformation, \textit{ARA} robot also collects steel surface images to performs visual inspection. The steel images are sent to a CNN network, which is segmented into corroded or healthy regions. A snapshot of the result extracted from the CNN network is shown in Fig. \ref{fig:mobileMove}(a).

For \textit{inch-worm transformation}, \textit{KDL} Inverse Kinematics and \textit{RRTConnect} motion planner in \textit{MoveIt} package are selected to implement the task, which calculates the robot inverse kinematics and generates a trajectory for an inch-worm jump from point $P_{conv}$ to the target plane. To do that, a primitive robot model is built in \textit{urdf} format, with the exact dimensions, joint types and limits to \textit{ARA} robot.  The generated trajectory - a ROS topic - is a set of robot joint angles, which the robot joints follow to reach the target pose. 

\begin{figure}[ht]
\centerline{\includegraphics[width=1\linewidth]{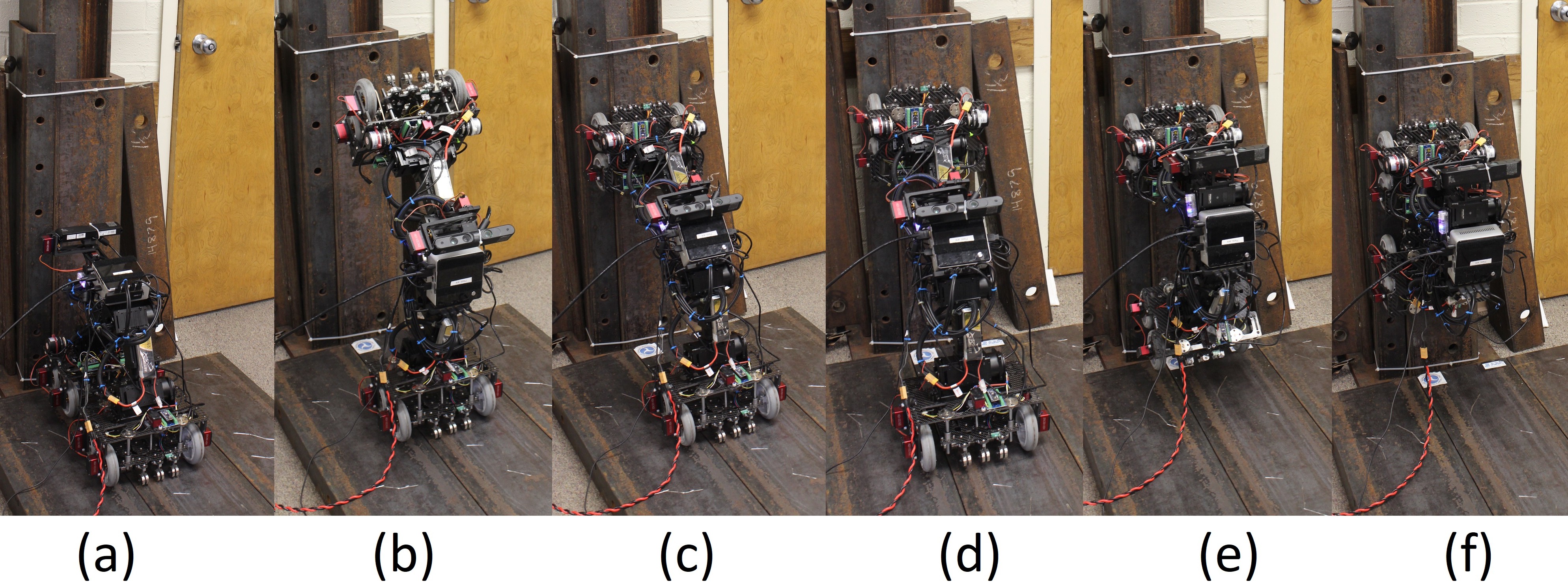}}
    \caption{\textit{inch-worm transformation}: a) magnetic array of second foot touches the base surface, b) first foot moves to convenient point, c) first foot reaches target pose and touches the second surface, d) magnetic array of second foot is released, e) and f) second foot moves to target pose}
    \label{fig:inch-wormjumpmotion}
\end{figure}

The inch-worm performance of the robot is shown in Fig. \ref{fig:inch-wormjumpmotion}. In the beginning, the robot activated and lowed down the magnetic array to touch the steel surface; then it transforms from the mobile configuration to the convenient pose $P_{conv}$ by following a predefined trajectory as shown in Fig. \ref{fig:inch-wormjumpmotion}(a) and Fig. \ref{fig:inch-wormjumpmotion}(b). As reaching point $P_{conv}$, the robot started following the \textit{RRTConnect} trajectory. As the first foot reached the target surface, the robot switched both magnetic arrays working modes, the one on the first foot was changed to \textit{touched mode}, and the other was set to \textit{untouched mode} as shown in Fig. \ref{fig:inch-wormjumpmotion}(c)-(d). Next, the second robot foot transforms into the target plane as shown in Fig. \ref{fig:inch-wormjumpmotion}(e)-(f). The whole robot operation was filmed, and the video-clip was uploaded at \url{ https://youtu.be/SHk5IIOBRdA} which was sped up three times than the experimental operation.  

\section{Conclusion and Future Work}\label{S.5}
A switching control mechanism for autonomous navigation of bridge-inspection robots is proposed in this work. The unique feature of switching in two modes enhances the flexibility of navigation and inspection. The most significant part of this research is the estimation of available non-convex surface for navigation using area, plane and height availability. Moreover, the mobile control framework and magnetic adherence distance controller proposed in this work are very important for robot navigation. Nonetheless, the integration of different parts of the framework into the real-world environment was the most challenging part of this research. The motion planner \textit{RRTConnect} was not robust in calculating the inverse kinematics and generating the trajectory for the robot, and needed to redo sometimes. Further investigation of deployment in actual steel bridges, building a new motion planner for this robot, and optimization of inch-worm transformation is necessary as the next phase of our research.

\bibliographystyle{unsrt}
\bibliography{RefFile}
\end{document}